\RequirePackage{amsmath}
\documentclass[runningheads,a4paper]{llncs}\usepackage[]{graphicx}\usepackage[]{color}
\makeatletter
\def\maxwidth{ %
  \ifdim\Gin@nat@width>\linewidth
    \linewidth
  \else
    \Gin@nat@width
  \fi
}
\makeatother

\definecolor{fgcolor}{rgb}{0.345, 0.345, 0.345}

\usepackage{framed}
\makeatletter
 {\par\unskip\endMakeFramed%
 \at@end@of@kframe}
\makeatother

\definecolor{shadecolor}{rgb}{.97, .97, .97}
\definecolor{messagecolor}{rgb}{0, 0, 0}
\definecolor{warningcolor}{rgb}{1, 0, 1}
\definecolor{errorcolor}{rgb}{1, 0, 0}
\newenvironment{knitrout}{}{} % an empty environment to be redefined in TeX

\usepackage{alltt}
\usepackage{microtype}
\usepackage{amsfonts}
\newcommand{\inlinecode}{\texttt}
\newcommand{\Xe}{\mathbf{X}_\textrm{e}}

\usepackage[%
rm={oldstyle=false,proportional=true},%
sf={oldstyle=false,proportional=true},%
tt={oldstyle=false,proportional=true,variable=true},%
qt=false%
]{cfr-lm}
\usepackage{graphicx}

\usepackage{paralist}
\usepackage[caption=false]{subfig}
\usepackage[T1]{fontenc}

\usepackage[math]{blindtext}

\usepackage{csquotes}

\usepackage{microtype}

\usepackage{url}
\makeatletter
\def\Url@twoslashes{\mathchar`\/\@ifnextchar/{\kern-.2em}{}}
\g@addto@macro\UrlSpecials{\do\/{\Url@twoslashes}}
\makeatother
\makeatletter
\g@addto@macro{\UrlBreaks}{\UrlOrds}
\makeatother

\usepackage[
bookmarks=false,
breaklinks=true,
colorlinks=true,
linkcolor=black,
citecolor=black,
urlcolor=black,
pdfpagelayout=SinglePage,
pdfstartview=Fit
]{hyperref}
\usepackage[all]{hypcap}

\usepackage[capitalise]{cleveref}
\crefname{section}{Sect.}{Sect.}
\Crefname{section}{Section}{Sections}
\crefname{figure}{Fig.}{Fig.}
\Crefname{figure}{Figure}{Figures}

\usepackage{xspace}
\DeclareFontFamily{U}{MnSymbolC}{}
\DeclareSymbolFont{MnSyC}{U}{MnSymbolC}{m}{n}
\DeclareFontShape{U}{MnSymbolC}{m}{n}{
    <-6>  MnSymbolC5
   <6-7>  MnSymbolC6
   <7-8>  MnSymbolC7
   <8-9>  MnSymbolC8
   <9-10> MnSymbolC9
  <10-12> MnSymbolC10
  <12->   MnSymbolC12%
}{}

\DeclareMathSymbol{\powerset}{\mathord}{MnSyC}{180}

\IfFileExists{upquote.sty}{\usepackage{upquote}}{}
\begin{document}

\title{Reproducible Pattern Recognition Research:\\The Case of Optimistic SSL}

\author{Jesse H. Krijthe\inst{1,2} \and Marco Loog\inst{1,3}}

\institute{
Pattern Recognition Laboratory, Delft University of Technology\\
\and
Department of Molecular Epidemiology, Leiden University Medical Center\\
\and
The Image Section, University of Copenhagen\\
\email{jkrijthe@gmail.com}
}
			
\maketitle

\begin{abstract}
In this paper, we discuss the approaches we took and trade-offs involved in making a paper on a conceptual topic in pattern recognition research fully reproducible. We discuss our definition of reproducibility, the tools used, how the analysis was set up, show some examples of alternative analyses the code enables and discuss our views on reproducibility.
\end{abstract}


\section{Introduction}
The goal of this work is to describe and discuss the choices involved in making the results of a conceptual work in pattern recognition fully reproducible. Conceptual, here, refers to the type and goal of the analysis that was done in that work: using simulations and experiments, it tries to improve our understanding of one or more methods, rather than apply an existing method to some new application or introduce supposedly novel approaches. The work in question is our paper on \textit{Optimistic Semi-supervised Least Squares Classification} \cite{Krijthe2016a}, which reports on two ways in which a supervised least squares classifier can be adapted to the semi-supervised setting, the connections between these two approaches and why one of these approaches often outperforms the other.

The conceptual nature of the work has particular advantages in making it reproducible: the data required to run experiments can easily be made available or, for simulated datasets, data are not required and the code to run the experiments is relatively self-contained, i.e. it has few dependencies on code outside this project. One could argue that for these types of projects, there is no reason \emph{not} to make results reproducible. We notice, however, that in practice, trade-offs and problems still come up. We will discuss our experience in this paper and use it as a case study to discuss the uses of reproducibility in pattern recognition research.

We will start by giving a short summary of the original paper on optimistic semi-supervised learning. We will then discuss what we mean by reproducibility and discuss the tools and strategies used here. After some examples of alternative analyses enabled by the reproducible nature of the work, we end with a discussion on the relevance of reproducibility in pattern recognition research.

\section{Summary of Optimistic SSL}
In supervised classification, classifiers are trained using a dataset of input/output pairs $\{(\mathbf{x}_i,y_i)\}^L_{i=1}$, where $\mathbf{x}_i$ is a $d$-dimensional input vector and $y_i$ is a binary outcome encoded using some value $m$ for one class and $n$ for the other. In semi-supervised learning, one attempts to use an additional set of unlabeled data $\{(\mathbf{x}_j)\}^U_{j=1}$ to improve the construction of a classifier to solve the supervised learning task. Semi-supervised learning is an active area of research due to its promise of improving classifiers in tasks where labeling objects is relatively expensive, or unlabeled data is inexpensive to come by.

The goal of the work in \cite{Krijthe2016a} is to study two different ways to adapt the supervised least squares classifier to the semi-supervised learning setting. The supervised least squares classifier for the two-class problem is defined as the linear classifier that minimizes the quadratic loss on the labeled objects or, equivalently, least squares regression applied to a numeric encoding of the labels, with the following objective function:
\begin{equation}
J_s(\mathbf{w}) = \| \mathbf{X} \mathbf{w}-\mathbf{y} \|^2 + \lambda \|\mathbf{w} \|^2 \,, \nonumber
\end{equation}
where $\mathbf{X}$ is the $L \times d$ design matrix of the labeled objects, $\mathbf{w}$ refers to the weights of the linear classifier and $\lambda$ is a regularization parameter. We now define two straightforward ways to include the unlabeled data in this objective function. The first we refer to as the \emph{label based objective}, since it treats the missing labels of the unlabeled data as a vector $\mathbf{u}$ that we should minimize over:
$$
J_l(\mathbf{w},\mathbf{u}) = \| \Xe \mathbf{w}-\begin{bmatrix} \mathbf{y} \\ \mathbf{u} \end{bmatrix} \|^2 + \lambda \|\mathbf{w} \|^2 \,,
$$
where $\Xe$ is an $(L+U) \times d$ design matrix containing the $d$ feature values for all, labeled and unlabeled, objects.
A second way to include the data is to consider that each unlabeled object belongs to one of two classes, and we can assign each object a responsibility: a probability of belonging to each class. If the classes are encoded as $m$ and $n$, for instance $-1$ and $+1$, this \emph{responsibility based objective} is defined as:
\begin{align}
J_r(\mathbf{w},\mathbf{q}) = & \| \mathbf{X} \mathbf{w}-\mathbf{y} \|^2 + \lambda \|\mathbf{w} \|^2 \nonumber + \sum_{j=1}^{U}  q_j (\mathbf{x}_j^\top \mathbf{w} - m)^2  + (1-q_j) (\mathbf{x}_j^\top \mathbf{w} - n)^2 \,. \nonumber
\end{align}

The first result from the paper is that applying block coordinate descent to these objectives -- where we alternate between minimizing over $\mathbf{w}$ and $\mathbf{u}$ respectively $\mathbf{q}$ -- the second procedure turns out to be equivalent to the well-known \emph{hard-label self-learning} approach applied to the least squares classifier, while the first approach is equivalent to a \emph{soft-label self-learning}, similar to a method that was originally proposed for regression as early as the 1930s \cite{Healy1956}.

The second result from the paper \cite{Krijthe2016a} is that the soft-label variant typically outperforms the hard-label variant on a set of benchmark datasets. In the paper we showed these results in terms of the error rate on an unseen test set: the learning curves of the performance for different amounts of unlabeled data are typically lower for the soft-label variant than for the hard-label variant. We will revisit these results in \cref{section:exampleperformance}, by showing how to adapt the code to not only consider the performance in terms of the error rate, but in terms of the quadratic loss used by the classifier as well.

The third result is a study of one reason for the performance difference by looking at the effect of local minima on the optimization problems posed by both approaches. We find that the label based objective corresponding to the soft-label variant has much fewer local minima for the optimization to get stuck in, compared to the hard-label variant, which often gets stuck in a bad local minimum, even though a better local minimum may be available.

\section{Reproducibility}

\subsection{Definition of reproducibility}
Reproducibility and replicability of experiments has gained increasing interest both in science in general \cite{Goodman2016a} and in pattern recognition/computer vision/machine learning as well \cite{Donoho2009,Drummond2009}. Much of this interest can be attributed to what some call the ``Reproducibility crisis'' in science: many published results can not be replicated by others trying to verify these results. Perhaps the most visible and laudable effort to estimate the scale of this problem in one scientific discipline has been the Open Science Collaboration's efforts in Psychology \cite{OpenScienceCollaboration2015} which finds that by some measures of replicability, the results of less than half of the $100$ studies selected for replication could actually be replicated. A related, but different phenomenon is the ``credibility crisis'' \cite{Donoho2009} which refers to the decrease in the believability in computational scientific results caused by the increasing difficulty to understand exactly how results were obtained based on the textual description alone.

While ``replicability is not reproducibility'' \cite{Drummond2009}, these terms on their own may already refer to different things. Reference \cite{Goodman2016a} attempts to give clear definitions for different notions of reproducing a result. In this paper, we are mostly concerned with what they call \emph{methods reproducibility}, meaning the ability of different researchers to reproduce exactly the same figures and tables of results based on the data, code and other artefacts provided by the original authors. Like \cite{Patil2016}, we will refer to this simply as \emph{reproducibility}. Note that the moniker reproducibility does not say anything about the correctness of results, only that they can be obtained again by a different researcher.

Also like \cite{Patil2016}, we will use the word \emph{replicability} to what \cite{Goodman2016a} calls \emph{results reproducibility}: the ability to obtain the results that support the same conclusion by an independent study. Here independent study is still vaguely defined to mean that we set up a new study, where we gather and analyse data using a procedure that ``closely resembles'' the procedures used in the original work.  This is what the ``reproducibility crisis'' we mentioned at the start of this section refers to: not being able to obtain the same results by such studies. In the pattern recognition context, this definition could often come down to the exact same thing as reproducibility. The definition in \cite{Patil2016} is slightly more explicit and considers a study to be a replication if the population, question, hypothesis, experimental design and the analysis plan remain fixed, but the analyst and the code, for instance, have been changed. For a proper definition of a replication in pattern recognition research, one aspect of a replication could be a re-implementation of methods. We will come back to this in the discussion.

The reason we attempt to be so explicit about our definitions here is that the meaning of the words reproducibility and replicability is sometimes interchanged by other authors. Note, for instance, that by our definitions, the reproducibility crisis is best referred to as the replicability crisis. Or consider \cite{Drummond2009} who refers to methods reproducibility as replicability, and uses reproducibility to mean obtaining the same result using an independent study. 

While our definition of reproducibility only concerns the reproduction of the results in the original paper, we will illustrate that having reproducible results reduces the friction to make small changes to the code to explore alternative analyses. This allows one to explore, for instance, how sensitive the results are to particular parameter choices made by the original authors, or whether the method also works for slightly different datasets. In other words, like in a replication, where many things are changed at once to see whether a result can still be obtained, these small changes teach us something about the robustness of the results.

Even if we stick to our definition of `reproduce exactly the results', there are several levels at which this can be interpreted for a pattern recognition study like ours. We could, for instance, consider the following levels:
\begin{itemize}
\item Final paper can be reproduced from the source text
\item Figures and tables can be generated from results of computations
\item Results of computations can be generated from experiment datasets
\item Experiment datasets can be generated from raw data 
\end{itemize}
All using steps for which open source code is available. Although we consider a paper reproducible when all these steps are fulfilled, in practice we will show that for many of the benefits of reproducibility, it may be useful to consider these as separate steps: to explore the effect of a different outcome measure, we may not want to redo the computations. Or for a particular experiment, the preprocessing applied to the raw data may not be particularly relevant, as long as we have the processed data.

\subsection{Strategy for Reproducibility}
All the code used to produce the results in \cite{Krijthe2016a} is written in the R programming language \cite{RCoreTeam2016}, while the paper itself is a combination of Latex and R code to generate the figures. The two are combined using the knitr package \cite{Xie2014}. knitr allows one to intersperse Latex with blocks of R code that get executed and turned into Latex expressions or figures before the Latex document is compiled. This allows for the code that generates the figures to be placed where one would usually place a figure environment in the Latex document, so that everything that visually becomes part of the paper is defined in a single document. One of the advantages of this approach is that the author can be sure that the figures and tables in the paper were actually generated by this code, i.e. the code did not inadvertently change in the meantime.

In principle, one could also include the code for the experiments itself in this document. We noticed, however, that even for projects of this relatively small size, and even though knitr is able to optionally cache results, we found it more convenient to place the code of the experiments in separate files, save the results to an R data file, and then load these result files to be used in the generation of the figures in the knitr document.

The advantage of this particular approach to splitting the computations across files was that we could easily transfer the experiment code to a compute server to run the experiments, while writing the document. A disadvantage of not including the experiment code in the final document is that it increases the possibility that the chain of reproducibility is broken: for instance, we could apply some transformation to the data between the time the experiments where run and the figures are generated and forget, or accidentally save or load an old result file.

Another trade-off was between writing code for this particular analysis project or splitting code off into separate packages. For instance, for the implementations of the classifiers, we decided to make these part of a larger package of methods for semi-supervised learning \cite{RSSL}. This makes the methods and some code used to run experiments available for other applications. It also made sense here, since this project was part of a larger research programme into semi-supervised learning. The downside is that it introduces dependencies between projects. The main practical lesson we learned here is to save the reference to the particular version that was used to generate the results in the paper in the version control system, so that future changes do not effect one's ability to reproduce the results.

Similar to the implementations of the methods, we split the code used to load the datasets into a separate project, to be used for other projects. These scripts download the data and save them locally, unless this is already done previously.

\section{Examples}
\label{section:exampleperformance}
In this section we will show some additional analyses that are possible by changing the original code from \cite{Krijthe2016a} and that lead to some additional insights into the methods covered in the paper. The examples shown here are meant to illustrate that reproducible results have utility beyond the mere fact that we are sure how the results were produced: it allows for small changes by readers that can lead to additional insights. We order the examples by the size of the changes to the code required to obtain the results.

\subsection{Changing an example figure}
We start with the simplest case where small changes to the code that generates a figure can help illustrate a point. In the original paper we give an example why the soft-label self-learning variant would update the decision boundary using the unlabeled objects, and that this updating depends on the location of the unlabeled objects. Here we change the location of the unlabeled objects, by changing the line \inlinecode{X\_u <-  matrix(c(-1, 4), 2, 1)} to \inlinecode{X\_u <-  matrix(c(-1, 0.5), 2, 1) } to show that when the decision values for all unlabeled objects are within $[-1,1]$, the soft-label self-learning is no different than the supervised solution. The result is shown in \cref{fig:additional-simple-example}. Note that this would be a case where an interactive version of the plot could be illustrative, instead of manually changing values and regenerating the plot.

\begin{knitrout}
\definecolor{shadecolor}{rgb}{0.969, 0.969, 0.969}\color{fgcolor}\begin{figure}

{\centering \includegraphics[width=0.8\linewidth]{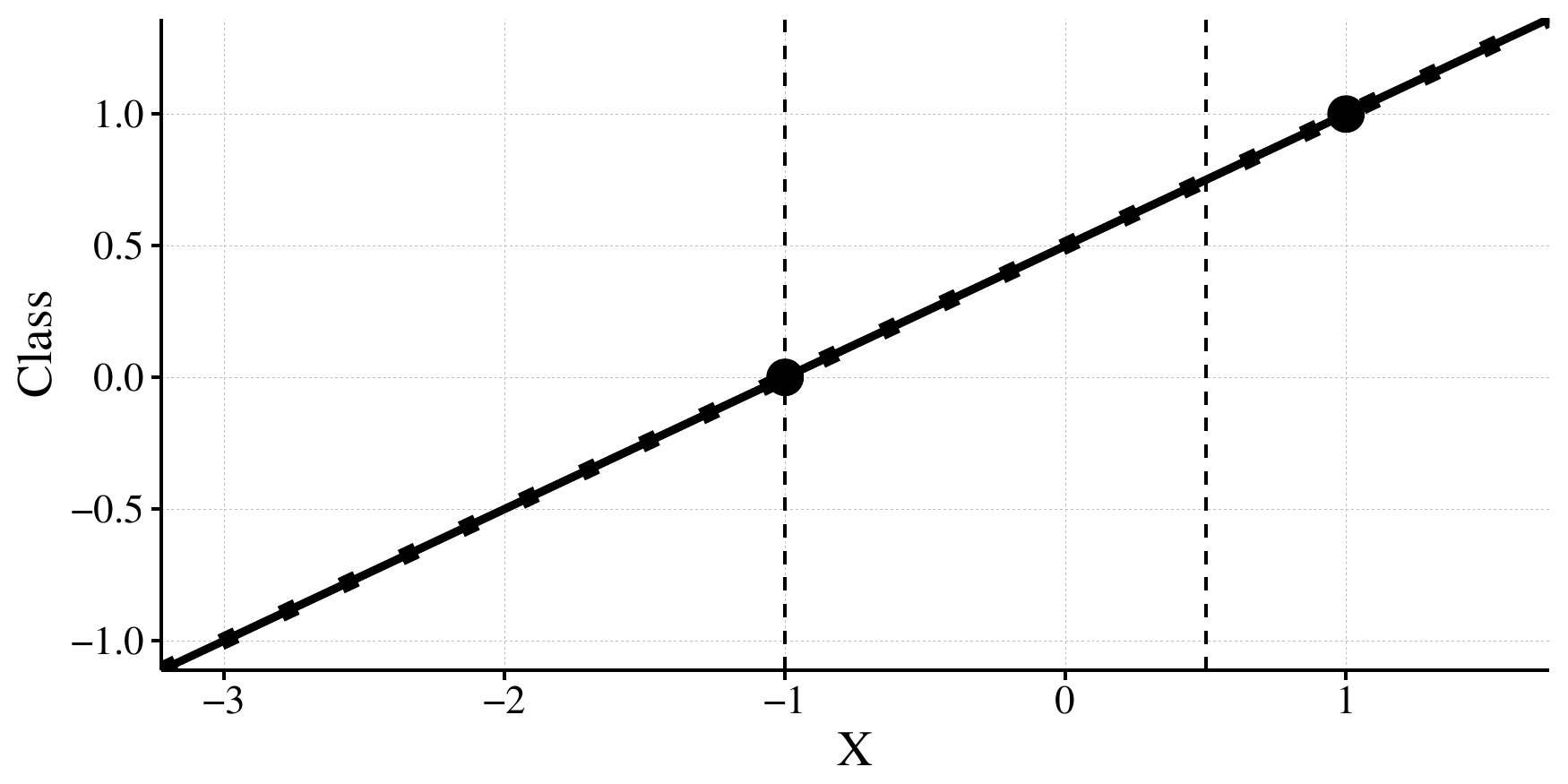} 

}

\caption[Example of the first step of soft-label self-learning]{Example of the first step of soft-label self-learning. The circles indicate two labeled objects, while the dashed vertical lines indicate the location of two unlabeled objects. The solid line is the supervised decision function. A dotted line indicates the updated decision function after finding the labels that minimize the loss of the supervised solution and using these labels as the labels for the unlabeled data in the next iteration. This last line is barely visible because the unlabeled data do not cause an update of the decision function in this case.}\label{fig:additional-simple-example}
\end{figure}

\end{knitrout}

\subsection{Changing the outcome quantity for the learning curves}
In the original paper, we report the error rate on a test set, for a fixed number of labeled training examples and an increasing amount of unlabeled examples. Alternatively, one might be interested in the performance in terms of the loss, instead of the classification error. Since this quantity was already computed during the experiment, we need not redo the experiment: a simple change in the code to plot the results suffices. More explicitly, we simply change the line \inlinecode{filter(Measure=="Error")} to \inlinecode{filter(Measure=="Average Loss Test")}.

\begin{knitrout}
\definecolor{shadecolor}{rgb}{0.969, 0.969, 0.969}\color{fgcolor}\begin{figure*}
\includegraphics[width=\maxwidth]{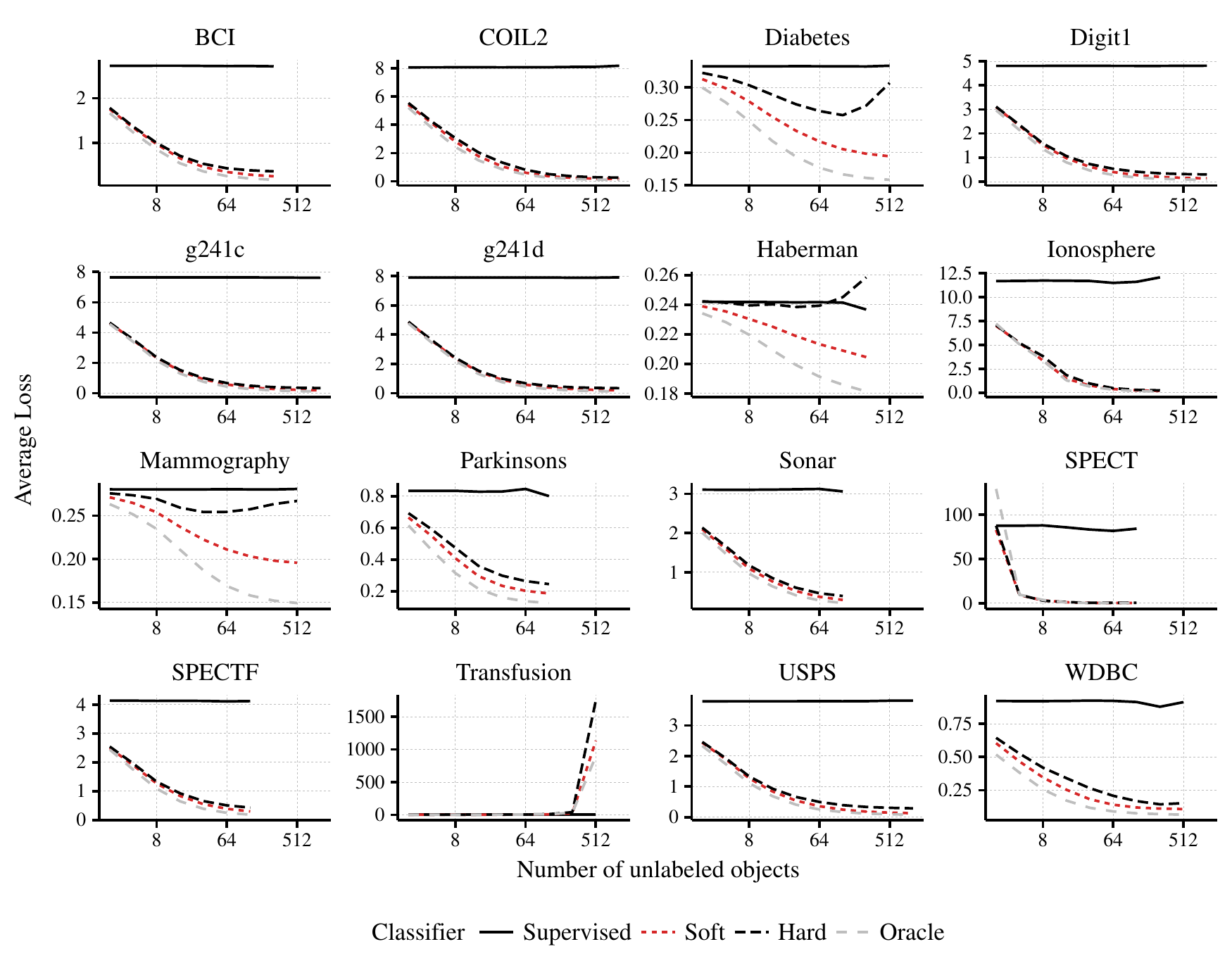} \caption[Average squared loss on the test set for increasing amounts of unlabeled training data]{Average squared loss on the test set for increasing amounts of unlabeled training data. The number of labeled objects remains fixed at a number larger than the dimensionality of the dataset to ensure the supervised solution is well-defined. Results are averaged over $1000$ repeats. Oracle refers to the supervised classifier that has access to the labels of all the objects.}\label{fig:learningcurves-loss}
\end{figure*}

\end{knitrout}

The results in \cref{fig:learningcurves-loss} show an interesting discrepancy when compared to the results in terms of the error rate: here in all cases the soft-label variant outperforms the hard-label variant, even on the dataset (Haberman) where it did not in terms of the error rate. Additionally, the loss starts increasing in more cases than for the error rate, especially for the hard-label variant.

\subsection{Sensitivity to random seed}
In the original work, we gave an example of a dataset where the hard-label self-learner is clearly outperformed by the soft-label self-learner. One might wonder how sensitive this dataset is to slight perturbations: is hard-label self-learning always much worse in this type of dataset or does it depend on the particular seed that was chosen when generating the data? This can be easily checked by changing the random seed and computing the classifiers. 

In \cref{fig:example-additional} we show two common configurations we find when we change the random seed. These configurations are qualitatively different from the result reported in the paper. In one case, there is no big difference between the two classifiers, unlike the result in the original work, while in the other, the hard-label self-learner gives deteriorated performance for a different reason: it assigns all objects to a single class. 

These results show that the original example is not stable to changes in the random seed. However, the conclusion that soft-label self-learning does not suffer from as severe a deterioration in performance as hard-label self-learning still holds. Our experience generating these additional examples does indicate, though, that other configurations than the prototypical example given in the paper are just as likely, if not or more likely, to occur.

\begin{knitrout}
\definecolor{shadecolor}{rgb}{0.969, 0.969, 0.969}\color{fgcolor}\begin{figure}
\subfloat[Small difference\label{fig:example-additional1}]{\includegraphics[width=.49\linewidth]{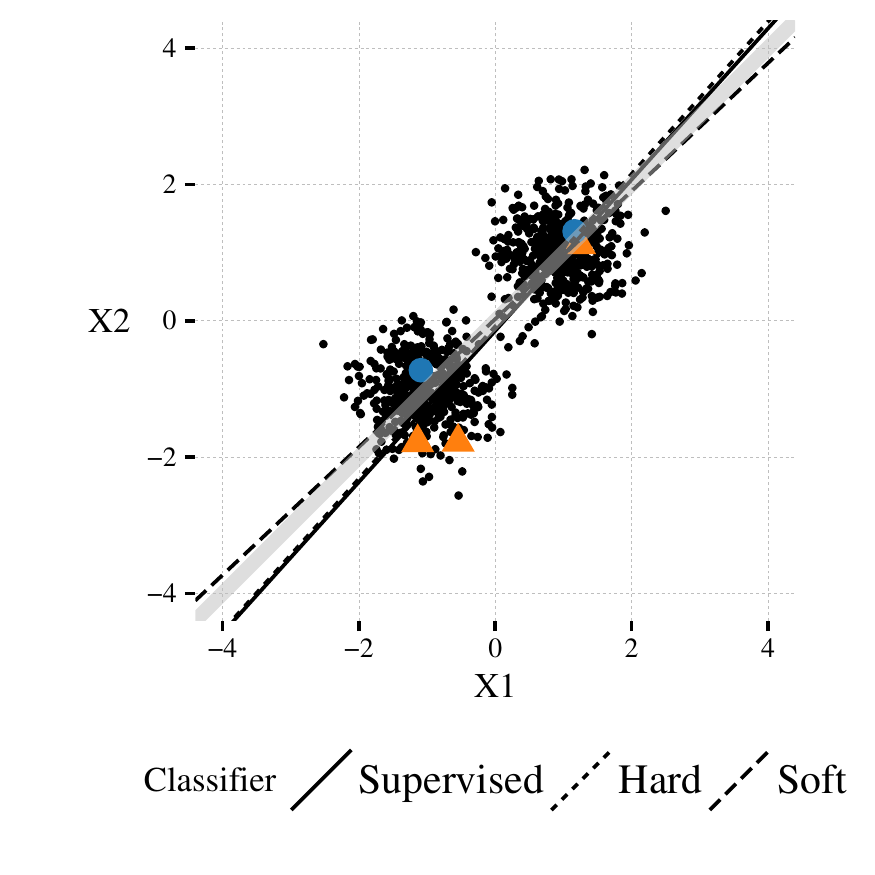} }
\subfloat[Hard-label single class\label{fig:example-additional2}]{\includegraphics[width=.49\linewidth]{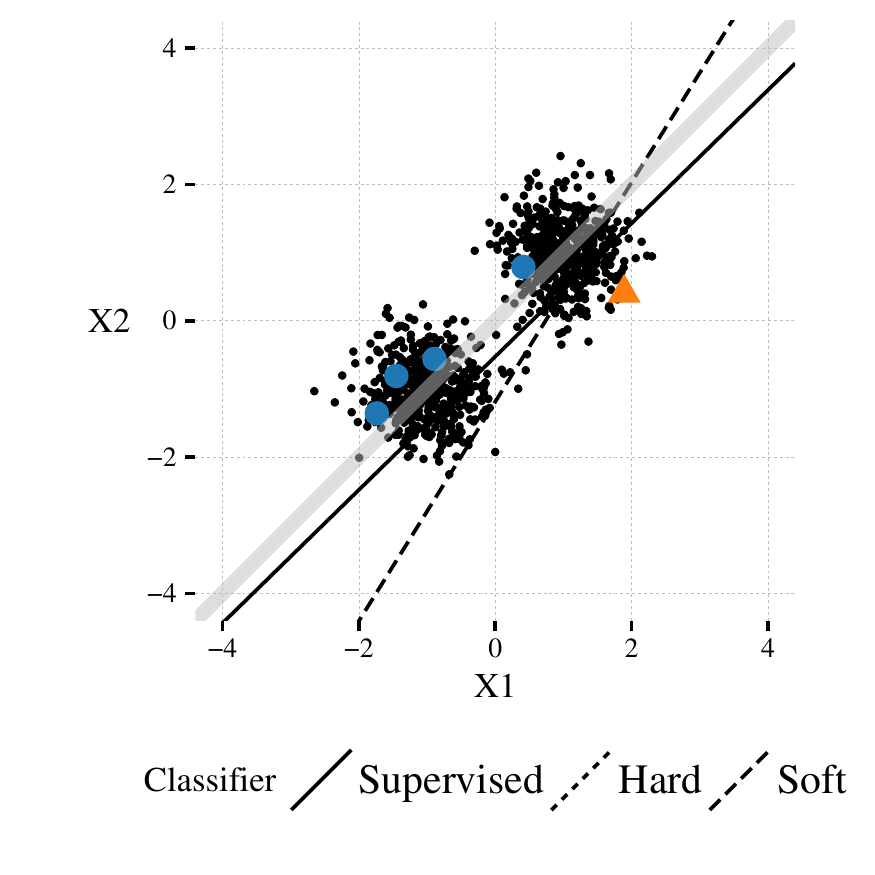} }\caption[Additional examples of the behaviour of hard-label and soft-label self-learning]{Additional examples of the behaviour of hard-label and soft-label self-learning. Light-grey line indicates true decision boundary. In (a), there is only a minor difference between soft-label and hard-label self-learning. In (b), the hard-label self-learner is not visible and assigns all objects to one class.}\label{fig:example-additional}
\end{figure}

\end{knitrout}

\subsection{Different type of learning curve}
For a more involved example, we use the code to generate a different type of learning curve. While we reported the learning curves for a fixed number of labeled samples and an increasing number of unlabeled samples, alternatively one could consider learning curves where the total number of training objects remains fixed, while the fraction of labeled objects is increased. Since the datasets are already available, we can easily set up these experiments by making some changes to the code that generates the other learning curves. We report these results in \cref{fig:learningcurves-frac}.

Although the ordering, in terms of performance, is similar in these curves as in the learning curves we originally reported, in many more cases the semi-supervised learners perform worse than the supervised learner. This indicates that as more labeled data becomes available, it is harder to outperform the supervised learner, especially since in these experiments, the amount of unlabeled data shrinks as we add more labeled data. Again, hard-label self-learning suffers more from degradation in performance than soft-label self-learning.

\begin{knitrout}
\definecolor{shadecolor}{rgb}{0.969, 0.969, 0.969}\color{fgcolor}\begin{figure*}
\includegraphics[width=\maxwidth]{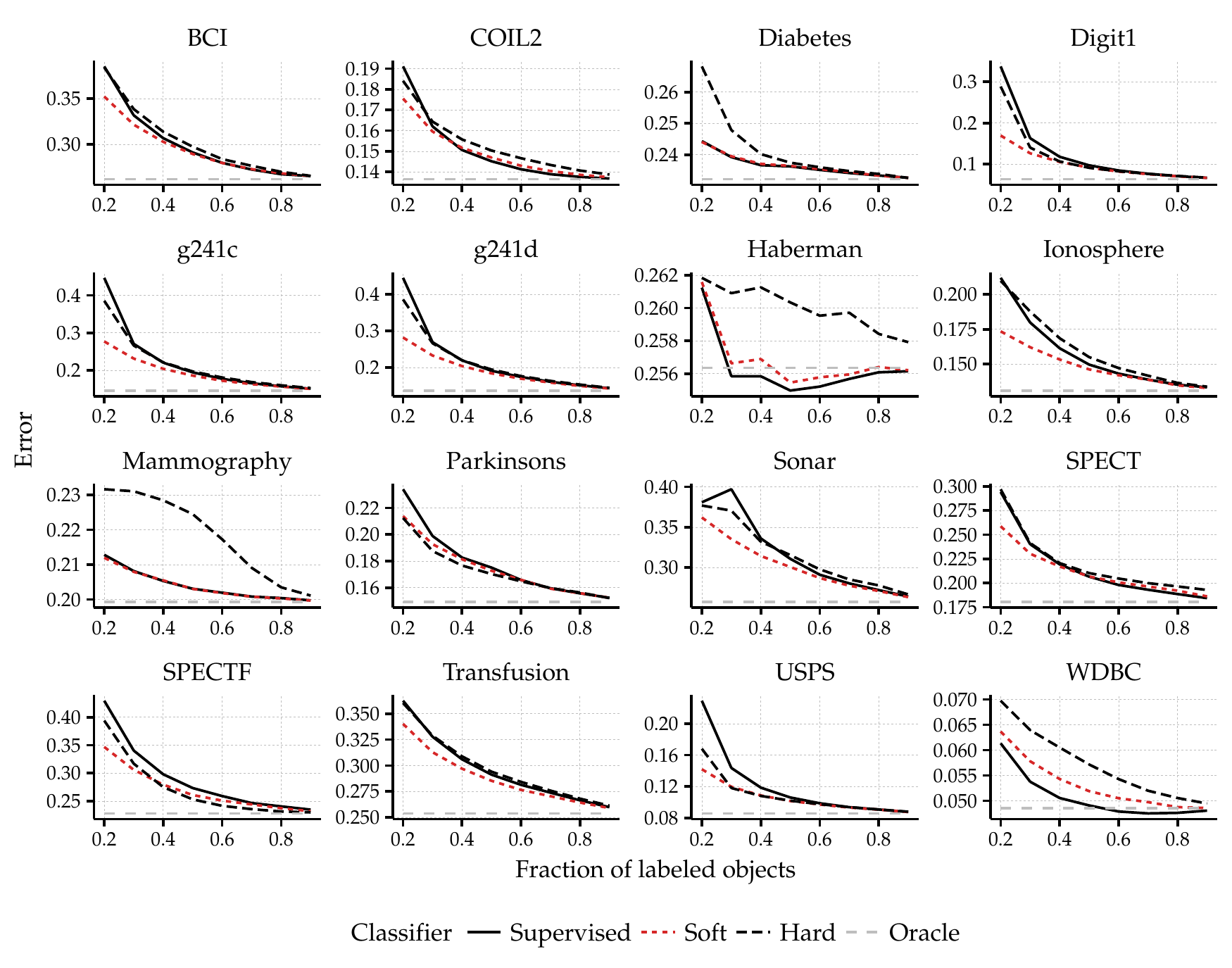} \caption[Classification accuracy when different fractions of the training set are labeled]{Classification accuracy when different fractions of the training set are labeled. $20\%$ of the data is left out as test data, the fractions indicate the fraction of objects of the remaining data that was labeled. Oracle refers to the supervised classifier that has access to the labels of all the objects.}\label{fig:learningcurves-frac}
\end{figure*}

\end{knitrout}

\section{Discussion}
While reproducible research is sometimes framed as being a requirement for a result to be believable \cite{Goodman2016a,Donoho2009}, we think it is important to emphasize that it does not just benefit scientific discourse, but has advantages for the researchers carrying out the original work as well. We elaborate in what follows.

\subsection{Advantages to the researcher}
Every research project is a collaboration. Sometimes with other individuals, but at the very least, a collaboration with yourself at some point in the future \cite[Ch. 13]{Wickham2015}. It is rare that one does not have to revisit results after they were originally generated. Making results reproducible ensures that collaborators and you yourself in the future can easily get back into old results and make changes.

Secondly, although reproducibility does not eliminate all errors, it makes it easier to catch some type of errors. For instance, errors introduced by copying and pasting results from one document to another. At the very least, it makes it easier to fix them.

On the whole, for the individual researcher, reproducibility reduces friction: it makes it easy to make changes to figures and experiments even after the whole analysis is done since the later steps in an analysis can be reused if they are implemented in a reproducible way.

\subsection{Advantages to scientific communication}

\subsubsection{The case for reproducibility}
Unlike the claim by \cite{Patil2016}, the requirement of reproducibility is not something ``everybody agrees'' on. In this respect, Drummond \cite{Drummond2009}  argues that replicating results is an important part of scientific progress, yet exactly reproducing results is a poor substitute that does not add much other than counter outright fraud, and reproducibility can become a distraction. It may, in other words, not be worthwhile to spend much resources on. 

This is perhaps a bit too pessimistic, for two reasons. 

First, while reproducibility says nothing about the correctness of a result, it does allow apparent mistakes to be more easily checked than if the code was not available. Consider, for instance, the well-known case of the finding of \cite{Herndon2014}, after much work, that the conclusions in a highly influential study on the effect of government debt on economic growth depended on a data coding error and were very sensitive to particular choices in the analysis. While reproducibility does not eliminate these errors, nor was it required to finally spot them, it would likely have sped up the efforts to uncover these errors. As this case shows, this can have real world consequences, since the original conclusions had been used as an argument around the world by proponents of austerity measures during the recent economic crisis.

Secondly, Drummond's main concern is that reproducibility only deals with keeping steps in an analysis pipeline fixed, while replicability is about changing things. However, as the case study in this paper has hopefully shown, an important side effect from exactly reproducing results is that it removes friction for both the original researchers and the community to make changes and build on the code. We have seen this has two advantages: it aids in communicating results and insights and it provides a stepping stone for others to build new results on.

\subsubsection{Replicability in Pattern Recognition}
One way to define replicability is to consider a study where the ``same procedures are followed but new data are collected'' \cite{Goodman2016a}, where this data is sampled from the same population. Is this definition of replicability then a useful construct in methodological pattern recognition research? In the pattern recognition context, data from the same population may be hard to define, if your population is a set of benchmark datasets. One could wonder whether results generalize to other problems. This however, does not fall under the conventional definition of replicability, but rather under the term \emph{generalizability}. 
In most sciences, one of the things we learn from a replication is what the essential conditions are that are necessary for a result to hold. Analogously, we argue one aspect of replicability in pattern recognition research is the implementation of the methods. As we have noticed in our own work, it is an under appreciated point how difficult or easy it is for another programmer or analyst to replicate the results of a method. It teaches us not just something about the competence of the programmer (a point that is often overstated) but also of the elegance of the method and its sensitivity to particular implementation choices that may have gone unnoticed and even unreported in the original work.

\subsubsection{Practicalities}
There is still a technical problem with reproducible results: how do we make sure they are still reproducible after programming languages, toolboxes and online platforms change or cease to exist? For centuries, the unit of the paper as the narrative artefact has proven to be a format that stands the test of time and changes in technology. In the work considered here, we refer to papers from the 1950s, which we where able to recover and which got its authors' point across perfectly well. We need to ensure this is still the case for the work produced today. The only proposal we have towards this is that, at the very least, the software used to produce results is produced using open source software. This both allows one to dig into every level of the implementation if this is required to answer a particular question, but also provides some chance of ensuring software is still available in a future where a particular software vendor may have ceased to exist.

\subsubsection{Going forward}
While we still consider reproducibility a worthwhile goal, there is a danger it leads to a false sense of security. Reproducibility is not replicability and it is replicability that constitutes progress in science. And reproducibility is not free: it requires effort on the part of the authors and reviewers of a manuscript. In the case covered in this paper, which is relatively easy to make reproducible, the advantages to the authors and the advantages to the community easily outweigh this effort. We should avoid dogmatism by realizing this trade-off might be different for other works.

\section{Conclusion}
We covered our approach to reproducing our paper on optimistic semi-supervised learning and showed some additional interesting, and nontrivial results by making slight adjustments to the figures and experiments which the reproducible nature of the paper allows. We argue that the advantages of reproducibility start during the research itself and extend to scientific communication. We need to realize, however, that reproducing results is not the same as replicating experiments, it primarily offers a poor but useful substitute.

\subsubsection*{Acknowledgement}
This work was funded by project P23 of the Dutch public/private research network COMMIT.

\bibliographystyle{splncs03}
\bibliography{library}
\end{document}